\documentclass[conference]{IEEEtran}

\usepackage{comment}
\usepackage{graphicx,subfigure,url,balance,comment} 
\usepackage{algorithmic} 
\usepackage{multirow} 
\usepackage{amsmath} 
\usepackage{amsfonts} 
\usepackage{epstopdf} 
\usepackage{caption2}
\usepackage{upquote}
\usepackage{tabularx}
\usepackage{authblk}
\usepackage{etoolbox}
\usepackage{color}
\usepackage{enumitem}
\usepackage[misc]{ifsym}
\usepackage[ruled]{algorithm2e}
\makeatletter
\patchcmd{\maketitle}{\@copyrightspace}{}{}{}
\makeatother

\usepackage{float}

\SetKwRepeat{doWhile}{do}{while}

\begin{document}

\title{A Data-driven Human Responsibility Management System}


\author{
  Xuejiao Tang$^1$, Jiong Qiu$^2$(\Letter), Ruijun Chen$^3$, Wenbin Zhang$^4$,\\
  Vasileios Iosifidis$^1$, Zhen Liu$^5$, Wei Meng$^6$, Mingli Zhang$^7$ and Ji Zhang$^8$ 
  \\
  $^1$Leibniz University Hannover, Germany $^2$Hangzhou Quanshi Software Co., Ltd, China \\
  $^3$National Cheng Kung University $^4$University of Maryland, Baltimore County, USA\\
  $^5$Guangdong Pharmaceutical University, China $^6$Beijing Forestry University, China\\
  $^7$McGill University, Canada $^8$University of Southern Queensland, Australia\\
  
  xuejiao.tang@stud.uni-hannover.de, colin\_qiu@hotmail.com, n78083016@mail.ncku.edu.tw, wenbinzhang@umbc.edu\\
  iosifidis@l3s.de, liu.zhen@gdpu.edu.cn, mnancy@bjfu.edu.cn, mingli.zhang@mcgill.ca, ji.zhang@usq.edu.au
  }

\maketitle 

\begin{abstract}
\label{sect:abstract}
An ideal safe workplace is described as a place where staffs fulfill responsibilities in a well-organized order, potential hazardous events are being monitored in real-time, as well as the number of accidents and relevant damages are minimized. However, occupational-related death and injury are still increasing and have been highly attended in the last decades due to the lack of comprehensive safety management. A smart safety management system is therefore urgently needed, in which the staffs are instructed to fulfill responsibilities as well as automating risk evaluations and alerting staffs and departments when needed. In this paper, a smart system for safety management in the workplace based on responsibility big data analysis and the internet of things (IoT) are proposed. The real world implementation and assessment demonstrate that the proposed systems have superior accountability performance and improve the responsibility fulfillment through real-time supervision and self-reminder.

\end{abstract}


\section{Introduction}
\label{sect:intro}
Improving the accountability and responsibility awareness has always been challenging in safety management~\cite{gerede2015study}. A large number of works have been proposed to avoid safety accidents through study the management methods or behavior-based system approaches~\cite{lee2016improvements,zhang2017phd,fernandez2007safety,zhang2017hybrid}. Wachter et al.~\cite{wachter2014system} propose a safety management practices and worker engagement system for reducing and preventing accidents, as well as investigating the significant relationships between safety system practices and accident rates. In~\cite{saracino2012new}, the M.I.M.O.S. (Methodology for the Implementation and Monitoring of Occupational Safety) is proposed to evaluate the safety management systems. The results of these studies have improved the safety management to some extent. However, heavily relying on human involvement has limited the effects of their systems as to err is human. On the contrary, data-driven artificial intelligence/machine learning systems with limited human involvement have become pervasive across all aspects of different domains~\cite{zhang2019faht,zhang2018deterministic,zhang2016using,zhang2019fairness}. A data-driven system managing responsibility with limited human involvement is also warranted. What's more, as Seok et al. mentioned~\cite{yoon2013effect}, there exist differences in occupational health and safety management system awareness in enterprises. Therefore, it is extremely difficult to execute the safety management completely correct~\cite{stolzer2015safety,zhang2018content}. In addition, the existing responsibility management mechanisms are ineffective in managing the increasing complexity of the hierarchy in enterprises~\cite{bovens1998quest,elmaraghy2012complexity,zhang2014comparison}. For instance, large enterprises with complex hierarchy ranging from leadership, supervision to security. The supervision of safety responsibility fulfillment has been weakened due to the need to decentralize other tasks, which would result in weak execution for safety responsibility~\cite{waddock2017total}. Last, timely discovering of risk problems is difficult when the management hierarchy is too complex~\cite{dobers2009corporate}. Consequently, a real-time and comprehensive supervising safety management system with limited human involvement is urgently needed~\cite{tang2019internet,chen2020internet}. 
 
To this end, this paper proposes a data-driven human responsibility management system. The proposed system is driven by responsibility big data with respect to risk pre-control, potential hazard, and accident handling, which is collected from the sensors and applications in IoT form as well as the submitted responsibility lists and responsibility fulfillment data by users. Analyzing the risk level and risk source of collected data, the dynamic knowledge-based responsibility sets of each unit (department) are defined. In addition, the directed acyclic graph (DAGs) is applied to sort out the relationship (supervise and be supervised, superior and subordinate, etc.) among responsibility nodes (an item in the responsibility list). Furthermore, the boundaries of responsibility used to clarify responsibilities of each unit (department) are automatically defined by combing the related responsibility sets of various positions. Besides, the proposed system reminds the staff to complete responsibilities according to the risk level or task cycle. Compared to human supervision, the proposed system aims to achieve higher management efficiency and real-time supervision.


\section{SYSTEM DESIGN}
\label{sect:design}
\begin{figure}[!h]
    \vspace{-3mm}
    \centering
    \includegraphics[width=0.9\linewidth, height=0.25\textheight]{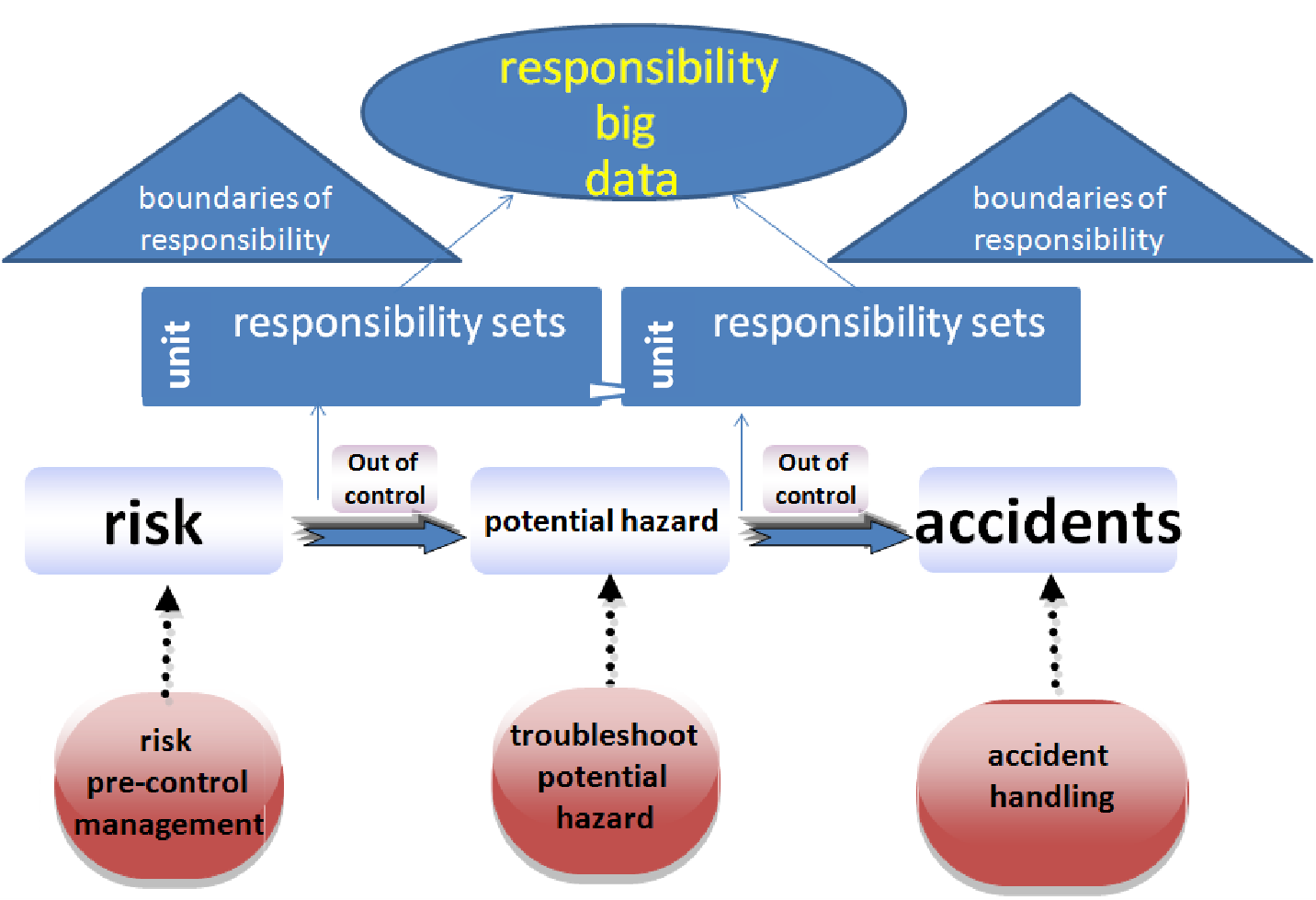}
    \vspace{-2mm}
    \caption{The structure of system.}
    \label{fig:figure1}
    \vspace{-2mm}
\end{figure}

Figure~\ref{fig:figure1} shows the proposed system structure. The data regarding risk, potential hazards, and accidents are collected using IoT or manually input. The safety management system analyzes the collected data and signals smart indicators providing in the system GUI to reflect the status of responsibility completion and ensuring a safe workplace.
The following subsection outlines the related definitions and key functions of the proposed system which enable the smart implementation of safety management.

\subsection{Related Definitions}
\begin{enumerate}[labelsep = .5em, leftmargin = 0pt, itemindent = 3em]
\item {\textbf{Boundaries of responsibility}} are defined to clarify the boundaries of accountability. They consist of the responsibility id, the original source of responsibility rules (regarding legal and other rules of safety management), the upper boundary of responsibility, the lower boundary of responsibility, classify rules of responsibility, relative cases.
\item {\textbf{Responsibility sets}} are the responsibility data sets of each unit (department) collected by IoT and the way that the staffs submit responsibility fulfillment data and set the necessary responsibility lists. Concretely, a responsibility set contains responsibility data of all positions in a related department. Moreover, a position is defined by various task items. Each task item is defined as a responsibility node. To better sort out the relationship among responsibility nodes, the DAGs are applied for a weighted relationship graph. The responsibility sets are defined by position id, superior position, subordinate position sets, primary responsibility category (major responsibility lists of each position), boundaries of responsibility, task type, trigger method, the hierarchy of responsibility category (subject responsibility, supervision responsibility, leadership responsibilities), senior position responsibility set, subordinate position responsibility set, risk id. 

\item {\textbf{Responsibility big data}} contains all of the input data along with processing data. Input data include IoT collected data and submitted data by users. Processing data such as data from boundaries of responsibility, responsibility attributes (completion progress, process evidence, regulatory evidence, consensus, assessment of completion), and responsibility sets (responsibility process data, responsibility coarseness calculation, relational superposition of directed acyclic graph, risk level analysis), are yield during system running.
\end{enumerate}

\subsection{Quantifying Responsibility}
Quantifying responsibility is an efficient way to clarify the responsibilities, supervisory responsibilities, and leadership responsibilities of relevant positions in various industry departments. First, the legal and professional responsibilities of various industries can be declared in the system by defining responsibility sets.
Next, the responsibility sets of each department will redefine the responsibility lists for each position. Sequentially, DAGs is applied among various task items which are added in responsibility lists to sort out the relationship among same level staffs, superior, and subordinate along with relationship among various departments. Besides, the responsibility lists with weight, task cycle, and evaluation rules information are precisely clarified to each position. Thus, each position obtains a set of qualifying responsibility lists. The responsibility lists are defined by different parameters (time, space, responsible region, etc). For different industries, especially the hazardous chemical industry, hazard monitoring and risk assessment are crucial for maintaining a safe workplace. The potential risks should be considered as a factor while designing the system so that responsibility lists contain the corresponding responsibility items. The detailed steps are as follows:


\begin{itemize}[leftmargin=*]
\item {\verb|Step 1|}: Create the responsibility sets and relevant boundaries of responsibility.


\item {\verb|Step 2|}: Sort out the hierarchy of the enterprises and identify the risk source and risk level of each position, then update the responsibility sets in Step 1 based on risk analysis of regarding positions.
\item{\verb|Step 3|}: Analyze the risk level in responsibility sets to automatically update the configuration of responsibility lists, such as responsibility cycle, trigger method, etc. (see Figure~\ref{fig:figure3}).
\item{\verb|Step 4|}: Create corresponding supervision responsibility lists based on the risk level of responsibility lists of each position.

\item{\verb|Step 5|}: Set the data collection method (by IoT sensors or manually input). 
\item{\verb|Step 6|}: Check if responsibility nodes (task items in responsibility lists) exists in the collected data. In case of a responsibility node is an item in collected responsibility data, link it to the relevant position. Otherwise, the system will connect them with referring responsibility lists of multiple positions.
\item{\verb|Step 7|}: Generate a configuration list of DAG (Directed Acyclic Graph)~\cite{colombo2012learning} for different levels of regulatory responsibility lists and subject responsibility lists.
\item{\verb|Step 8|}: Check responsibility performance status of positions with respect to risk level. In case the performance status cannot meet requirements, back to Step 3 - Step 7.
\end{itemize}

\subsection{Responsibility Evaluation and Supervision}

This stage refers to the submission of responsibility fulfillment and supervision data for the scoring process. Two components, the Internet of Responsibilities and the Internet of Things, are involved herein. The responsibility data and IoT~\cite{gubbi2013internet} data, which reflect the responsibility execution and machine running status, respectively, are collected by the proposed system. Thus, the fulfillment score will be evaluated, based on the status of responsibility fulfillment. Besides, the self-reminder module supervises the completion status of responsibility and reminds the staff to complete tasks on deadline or at a high risky level that can result in accidents. The self-reminder works as follows: 

\begin{itemize}[leftmargin=*]
\item {\verb|Step 1|}: List all task items should be reminded.
 \item {\verb|Step 2|}: Check each task item. Identify its attribute such as non-periodic and periodic tasks.
\item{\verb|Step 3|}: In case of a non-periodic task, check whether the task data is valid. If valid, sort the task into the list sets without self-reminders. Otherwise, sort the tasks into the reminder list sets.
\item{\verb|Step 4|}: In case of the periodic task, check the executed status, score status, and supervision status. If the execution is complete and the score is evaluated, sort it into the list that does not need to be reminded. Otherwise, add the tasks in the reminder list set and ranked them by the score.
\item{\verb|Step 5|}: For the reminder list sets, check the expiration time of each task then prioritize by the expiration time.
\end{itemize}

The evaluation process is detailed as below:
\begin{itemize}[leftmargin=*]
\item {\verb|Step 1|}: Obtain data from the submitting by user or IoT applications automatically.
\item {\verb|Step 2|}: Check the type of responsibility lists, if it is mandatory, check the current status.
\item{\verb|Step 3|}: If the current status can not result in risky accidents, score the corresponding tasks directly.
\item{\verb|Step 4|}: Check if there are other extension works and explain their benefits. The staffs can self-evaluate the corresponding score, and the score will be a part of the final results.
\item{\verb|Step 5|}: Check whether existing any supervisory responsibility, if there is supervision responsibility, submit supervision fulfillment data.
\item{\verb|Step 6|}: When the supervision data are submitted, the result of the supervisory is also considered for scoring. If the supervisory score is less than 50 percent of overall supervisory scores, the corresponding task completion score will be deducted in the same period.
\item{\verb|Step 7|}: Summarize the corresponding scores to get the recommended scores for each responsibility list.
\item{\verb|Step 8|}: The real-time responsibility score is evaluated during the execution.
\end{itemize}

\subsection{Scoring and Reporting}
The responsibility score of personnel will be reported weekly or monthly. The staffs with low responsibility scores will be reminded automatically. And the relevant departments will also be informed. The overall workflow is shown below:

\begin{itemize}[leftmargin=*]
\item {\verb|Step 1|}: Analyze the current incident.
\item {\verb|Step 2|}: List responsibility sets for accountability based on the responsibility lists and responsibility boundaries.
\item{\verb|Step 3|}: Determine the necessary positions of the enterprises based on the assessments of incidents.
\item{\verb|Step 4|}: List all the responsibility sets relevant to all of the responsibility data.
\item{\verb|Step 5|}: List positions with relatively low scores.
\item{\verb|Step 6|}: Analyze the correlation between the positions and responsibility lists among the staffs with relatively low scores.
\item{\verb|Step 7|}: Rank personnel with low scores and start the process of holding accountable.
\end{itemize}

\section{PRACTICAL APPLICATION}
\label{sect:experiment}
This paper has demonstrated the proposed system through a specific use case in a shopping mall. The responsibility lists and the completion status of responsibilities are visualized in the GUI interface. The type and evaluation method of responsibilities can be set in the system (see Figure~\ref{fig:figure3} and Figure~\ref{fig:figure4}). Also, an analysis result with weighted information will be given by the system to help improve safety management.
\subsection{Application for Quantifying Responsibility}
\begin{figure}[h]
    \vspace{-3mm}
    \centering
    \includegraphics[width=1\linewidth]{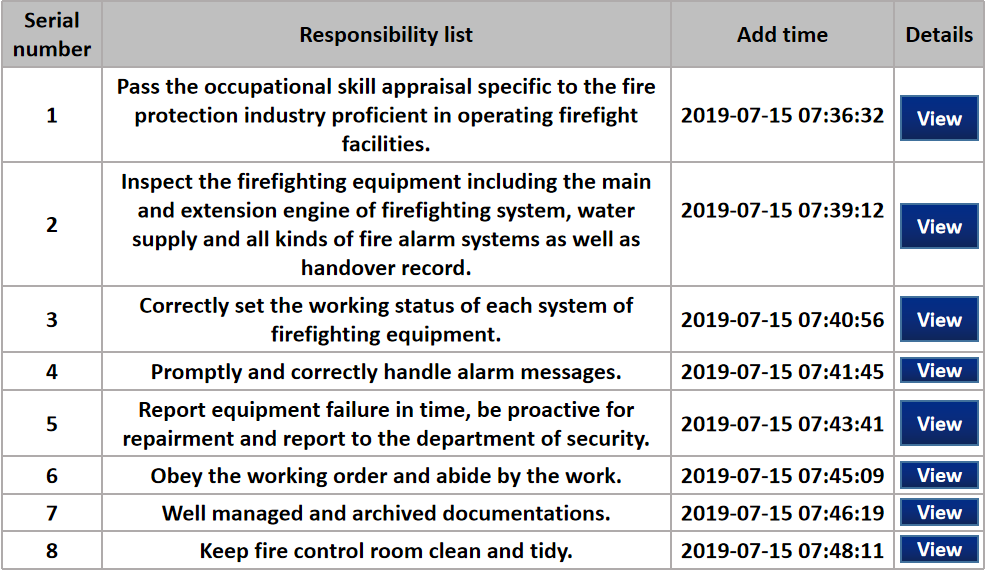}
    \vspace{-6mm}
    \caption{The responsibility lists of duty officer.}
    \vspace{-2mm}
    \label{fig:figure2}
\end{figure}
\begin{figure}[!h]
    \centering
    \includegraphics[width=0.9\linewidth]{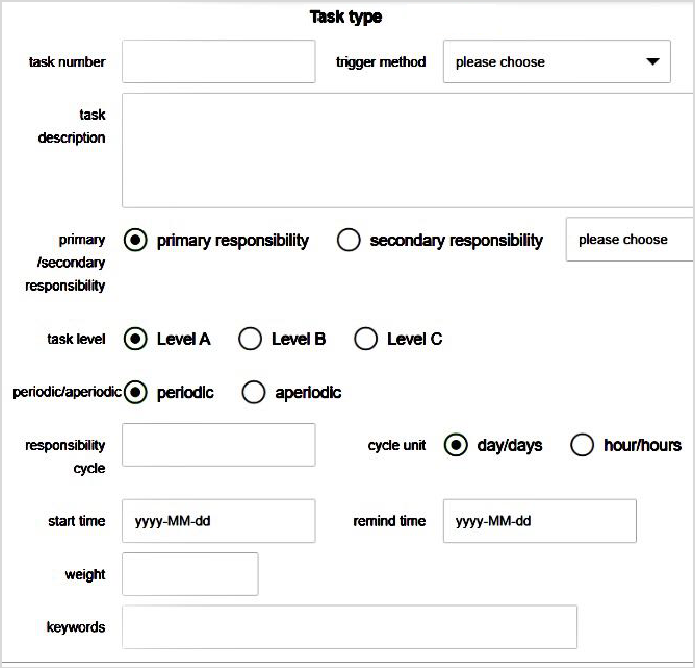}
    \vspace{-2mm}
    \caption{The task type configuration of responsibility list.}
    \label{fig:figure3}
    \vspace{-2mm}
\end{figure}
\begin{figure}[!h]
    \centering
    \includegraphics[width=0.9\linewidth]{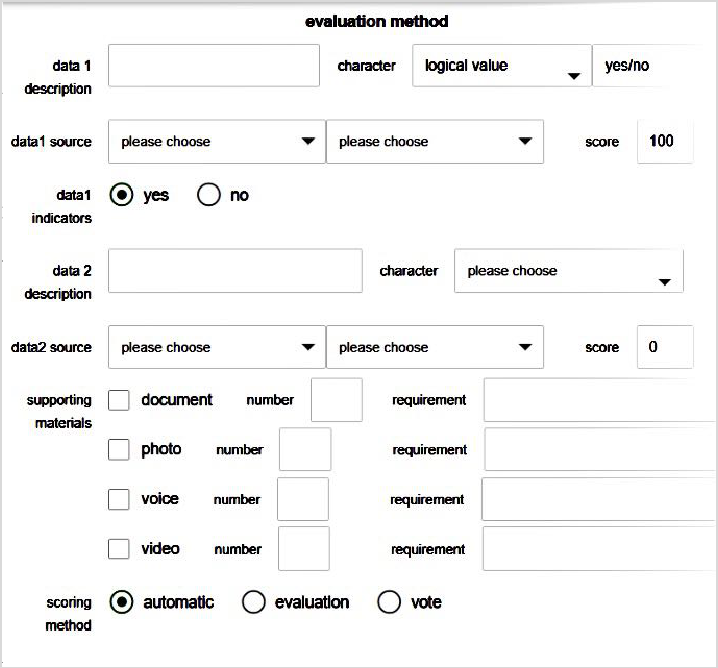}
    \vspace{-1mm}
    \caption{The evaluation method configuration of responsibility list.}
    \label{fig:figure4}
    \vspace{-5mm}
\end{figure}

With the specified responsibility list of different positions, the responsibility lists of duty officer are shown in Figure~\ref{fig:figure2}. Clicking the ``View'' button of a specific task in each row, the items of the selected responsibility list will be shown. Further clicks each responsibility list item, the configuration will be shown in Figure~\ref{fig:figure3} and Figure~\ref{fig:figure4}, where the attributes of the task type and evaluation method can be configured manually, such as setting the primary responsibility or the secondary responsibility. 
In addition, the risk level and cycle type (periodic or aperiodic) are listed in the items. If the task is periodic, the periodical information with the minimum cycle and the score weight can be set. The source data ( such as photos, documents, audio (sound recordings), videos, etc.) can be collected by IoT or manual input. Besides, the responsibility can be scored automatically, by evaluation or by voting. 

\subsection{Evaluation and Supervision Process}
\begin{figure}[!h]
    \centering
    \includegraphics[width=0.65\linewidth]{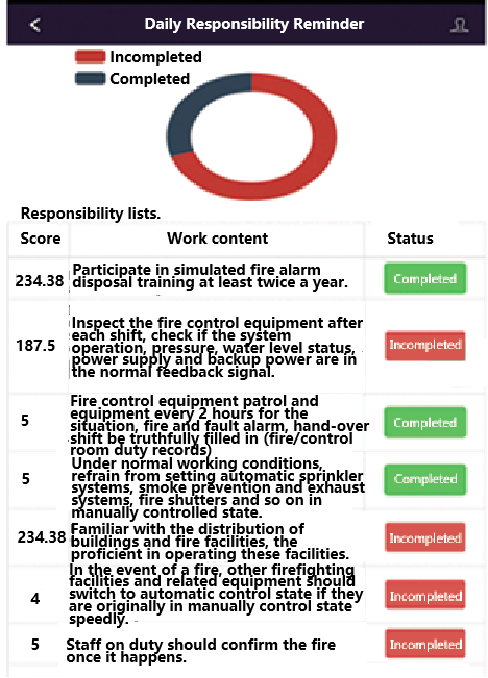}
    \vspace{-2mm}
    \caption{Completion status of responsibility lists.}
    \label{fig:figure5}
\end{figure}

\begin{figure}[!h]
    \centering
    \includegraphics[width=0.65\linewidth,height=0.3\textheight]{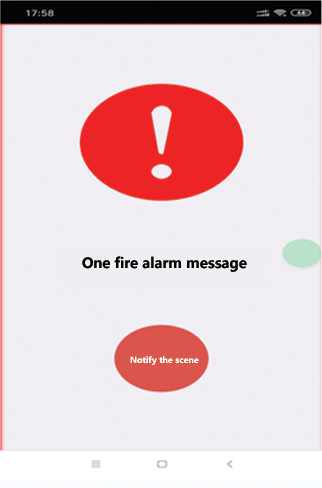}
    \vspace{-2mm}
    \caption{A fire alarm.}
    \label{fig:figure6}
    \vspace{-2mm}
\end{figure}
Figure~\ref{fig:figure5} shows the completion status of responsibility list. The incomplete items will be sorted according to the urgency degree and the weight of the task. Process data can be obtained from the IoT module. Figure~\ref{fig:figure6} shows a fire alarm example, duty officer informs the patrols once the fire alarms go off. After patrols are notified successfully, notification data with processing start and end time will be submitted in the system.

\subsection{Quantifying Accountability}
\begin{figure}[!h]
    \centering
    \includegraphics[width=0.9\linewidth,height=4.2cm]{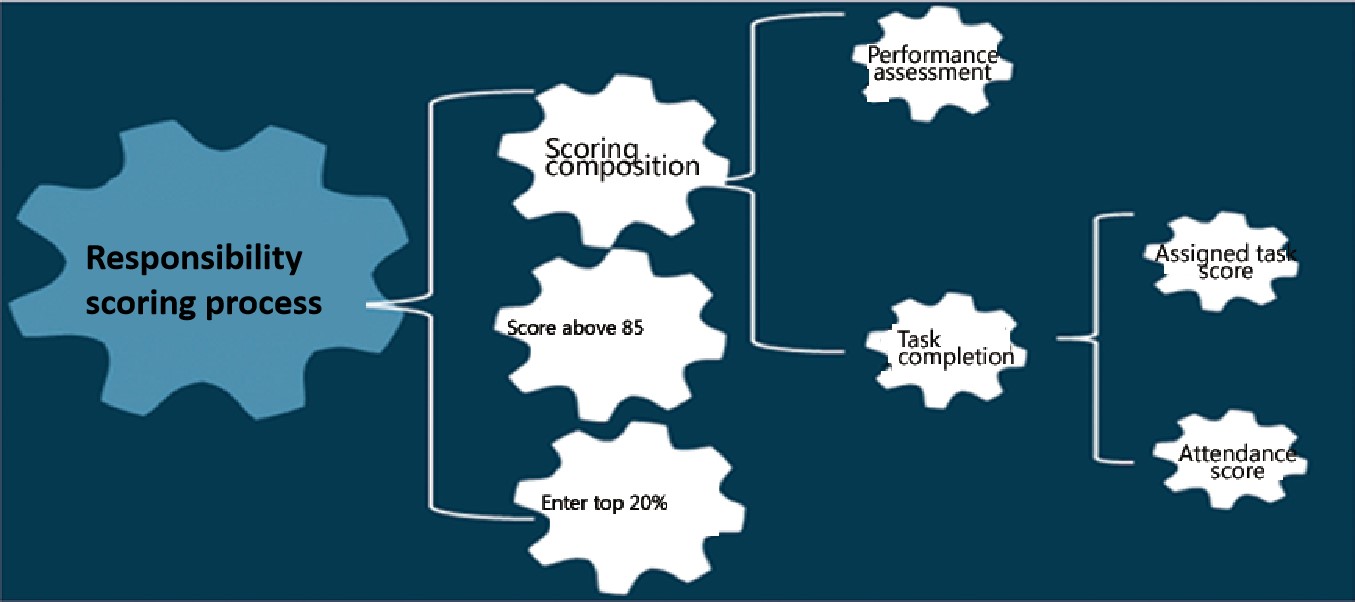}
    \caption{The scoring process.}
    \label{fig:figure7}
    \vspace{-2mm}
\end{figure}

As shown in Figure~\ref{fig:figure7}, the scoring composition is evaluated by measuring task completion and performance assessment. Besides, the task completion status is determined by assigned task scores and attendance scores, which could quantify the performance of responsibilities for each position.

\begin{figure}[!h]
    \vspace{-3mm}
    \centering
    \includegraphics[width=0.9\linewidth,height=4.5cm]{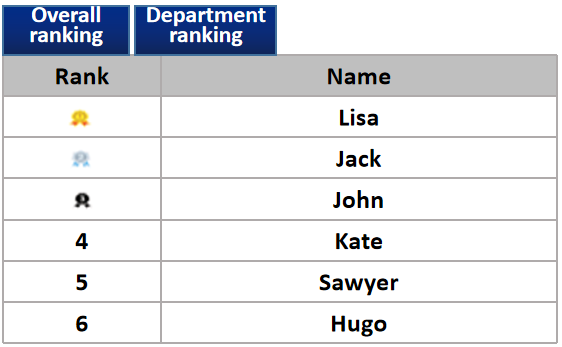}
    \vspace{-2mm}
    \caption{The score ranking.}
    \label{fig:figure8}
    \vspace{-4mm}
\end{figure}

As shown in Figure~\ref{fig:figure8}, personal responsibility scores can be sorted by overall and department ranking. The ranking results could be one of the performance evaluation basis.

\begin{figure}[!h]
    \centering
    \includegraphics[width=0.9\linewidth, height=0.2\textheight]{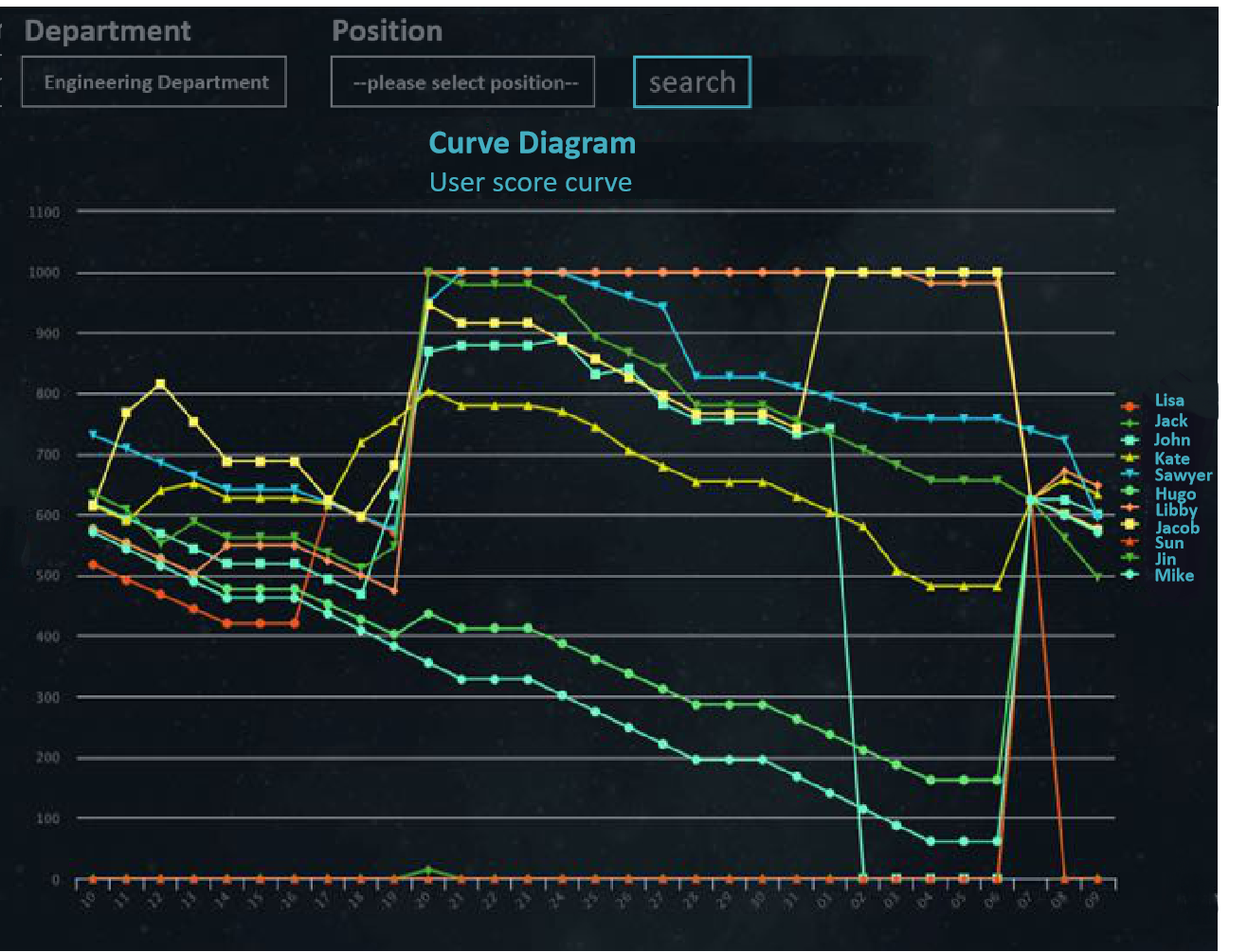}
    \vspace{-2mm}
    \caption{Personnel’s responsibility score curves.}
    \label{fig:figure9}
\end{figure}

\begin{figure}[!h]
    \centering
    \includegraphics[width=0.9\linewidth]{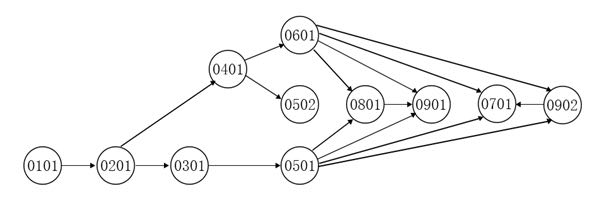}
    \vspace{-2mm}
    \caption{DAGs with responsibility weight information.}
    \label{fig:figure10}
    \vspace{-2mm}
\end{figure}

Figure~\ref{fig:figure9} shows the personnel's responsibility scores, which would change by time. The scores are evaluated in various aspects which intuitively reflect the effect and timeliness of responsibility completion.
Furthermore, the DAGs~\cite{vanderweele2007directed} graph with responsibility weight information in Figure~\ref{fig:figure10} sorts out the relationship between positions and responsibility nodes, which can improve the efficiency and accuracy of accountability. Based on the weight information, the status of responsibilities execution can be measured and visualized. 
\vspace{-5mm}

\section{Conclusion}
\label{sect:conclusion}
The proposed system aims at improving safety management in enterprises. The practical applications show its performance in various aspects, such as collecting data by using IoT, analyzing responsibility data and accountability items from multiple perspectives (risk pre-control, troubleshoot potential hazards, accident handing), visualization of the evaluation score, and supervision the responsibility fulfillment. In the practical application in a shopping mall, the proposed system visualizes the responsibility lists and the completion status of responsibility lists. Also, in case of an emergency such as a fire, the system will automatically alarm and send messages to notify the staff, so that the problems can be solved timely. In this way, the probability of accidents is greatly reduced. Moreover, the system evaluates the responsibility score by responsibility data and clarifies the results by visualized ranking results and curve diagram. Furthermore, the DAGs with responsibility weight information can clearly reflect the problems in safety management.

\bibliographystyle{IEEEtran}
\bibliography{ref}

\end{document}